# Marker-Based 3D Reconstruction of Aggregates with a Comparative Analysis of 2D and 3D Morphologies

**Haohang Huang[1]**
Graduate Research Assistant
Email: hhuang81@illinois.edu

**Jiayi Luo[1]**
Graduate Research Assistant
Email: jiayil5@illinois.edu

**Issam Qamhia[1], Ph.D.**
Postdoctoral Research Associate
Email: qamhia2@illinois.edu

**Erol Tutumluer[1], Ph.D.**
Abel Bliss Professor, Paul F. Kent Endowed Faculty Scholar
Email: tutumlue@illinois.edu

[1]University of Illinois at Urbana-Champaign
Department of Civil and Environmental Engineering
205 North Mathews, Urbana, Illinois 61801

**John M. Hart**
Principal Research Engineer, Computer Vision and Robotics Laboratory
University of Illinois at Urbana-Champaign
Beckman Institute for Advanced Science and Technology and
Coordinated Science Laboratory
1308 West Main Street, Urbana, Illinois 61801
Email: jmhart3@illinois.edu

**Andrew J. Stolba, P.G.**
Chief Geologist, Illinois Department of Transportation
126 E. Ash Street, Springfield, Illinois 62704
Email: Andrew.Stolba@illinois.gov

**ABSTRACT**

Aggregates, serving as the main skeleton in assemblies of construction materials, are important functional components in various building and transportation infrastructures. They can be used in unbound layer applications, e.g. pavement base and railroad ballast, bound applications of cement concrete and asphalt concrete, and as riprap and large-sized primary crushed rocks. Information on the size and shape or morphology of aggregates can greatly facilitate the Quality Assurance/Quality Control (QA/QC) process by providing insights of aggregate behavior during composition and packing. A full 3D characterization of aggregate particle morphology is difficult both during production in a quarry and at a construction site. Many aggregate imaging approaches have been developed to quantify the particle morphology by computer vision, including 2D image-based approaches that analyze particle silhouettes and 3D scanning-based methods that require expensive devices such as 3D laser scanners or X-Ray Computed Tomography (CT) equipment. This paper presents a flexible and cost-effective photogrammetry-based approach for the 3D reconstruction of aggregate particles. The proposed approach follows a marker-based design that enables background suppression, point cloud stitching, and scale referencing to obtain high-quality aggregate models. The accuracy of the reconstruction results was validated against ground-truth for selected aggregate samples. Comparative analyses were conducted on 2D and 3D morphological properties of the selected samples. Significant differences were found between the 2D and 3D statistics. Based on the presented approach, 3D shape information of aggregates can be obtained easily and at a low cost, thus allowing convenient aggregate inspection, data collection, and 3D morphological analysis.

## INTRODUCTION AND BACKGROUND

Aggregate materials serve as a very important functional component in the construction industry, and their shape or morphological properties demonstrate primary influence on performance of aggregate assemblies. Small-sized aggregates enhance the composite strength by filling the void in an aggregate mixture or by binding with other materials to form a matrix in mixtures, such as in asphalt concrete and Portland cement concrete (*1, 2*). Relatively larger aggregates, such as those in unbound pavement layers and railway ballast, provide efficient load bearing and distribution by particle interlocking (*3, 4, 5, 6, 7*). At even larger sizes, the medium-sized and large-sized aggregates are typically used as riprap materials for water/ice erosion control in hydraulic applications (*8, 9*). For all size groups of aggregates, the information on aggregate morphology can greatly assist the Quality Assurance/Quality Control (QA/QC) process by providing insights of aggregate behavior during composition and packing.

Imaging approaches that can reproduce aggregate particles in digital forms have been widely used in research applications as a quantitative way of analyzing aggregate morphology. Many successful 2D image-based approaches have been developed to capture the 2D morphological properties of aggregates. Masad et al. (*10*) and Gates et al. (*11*) developed the Aggregate Imaging System (AIMS) that can capture 2D particle shapes of aggregates spread onto a flat surface. Tutumluer et al. (*12*) and Moaveni et al. (*13*) developed the Enhanced-University of Illinois Aggregate Image Analyzer (E-UIAIA) which analyzes aggregate shape from three orthogonal views when passing over a conveyor belt. For relatively large-sized aggregates, Huang et al. recently designed an imaging system for the convenient volumetric estimation of aggregates under field condition (*14*) and developed a morphological analysis software, I-RIPRAP, for stockpile aggregate image analysis (*15, 16*). Zheng and Hryciw (*17*) conducted morphological analysis on soil grains from a two-view stereo photography analysis. All these approaches entail either pure 2D image analysis, or 2D analysis with limited reconstruction or approximation of the 3D surface. Thus, there are several limitations and assumptions for obtaining 3D aggregate models using these 2D image-based approaches.

To fully reconstruct the aggregates as 3D models, many 3D scanning-based approaches have been developed in the past decade. Anochie-Boateng et al. (*18*) and Komba et al. (*19*) designed and developed a 3D laser scanning device to obtain 3D aggregate models by a spot-beam triangulation scanning method. Jin et al. (*20*) constructed 3D solid models of nine aggregates by merging X-Ray Computed Tomography (CT) slices from the cross sections of the specimens. Complicated searching and merging algorithms were developed to orient the CT slices to form valid 3D shapes. Thilakarathna et al. (*21*) used a structured light 3D scanner to reconstruct 3D models by projecting preset light patterns onto the aggregate surface. However, these 3D scanning-based approaches usually utilize expensive scanning devices and require external lighting sources. Alternatively, more convenient and cost-effective photogrammetry approaches were investigated and demonstrated a comparable reconstruction quality when compared to the approaches requiring expensive imaging devices. Paixão et al. (*22*) reconstructed 18 ballast particles by fixing the aggregate with a support pedestal and obtaining all-around views at three elevations. The particle sizes were below 7.6 cm (3 in.) to ensure stable support from the pedestal. The photogrammetry results were compared with the results from 3D laser scanning, and both methods demonstrated very close results. Ozturk et al. (*23*) followed a similar photogrammetry procedure that captures all-around views from different viewing angles when the aggregate particle is glued to a stick and elevated in the air. The particles sizes were around 1.3 cm (0.5 in.) to be stably fixed using glue. Both researchers used a support system to elevate the aggregate in the air such that all-around views are accessible. However, the size range of aggregates that can be reconstructed by the procedure is greatly limited by the design of the support system.

Based on the literature review of available techniques, the major limitations of existing aggregate reconstruction systems are as follows:

- *Devices are costly*. Most of the aggregate imaging systems that can obtain high-fidelity 3D aggregate models involve expensive devices such as 3D structured light scanner, 3D laser scanner, or X-Ray CT scanner. Commercial software tools usually come with the expensive devices. Photogrammetry-based methods using digital cameras have a much lower cost but may lack a well-established pipeline for the pre-processing and post-processing of the data.
- *Limited range of aggregate sizes that can be scanned*. Unlike X-Ray CT devices, 3D laser scanners and 3D structured light scanners can generally scan a larger size range of aggregates. For existing photogrammetry-based approaches, however, the feasible size ranges are greatly limited because the procedure uses a support system to elevate the aggregate in the air for all-around inspection.
- *Operating condition.* The available 3D systems require sophisticated light control, especially for the 3D structured light devices. Photogrammetry-based approaches have more relaxed restrictions on the operating condition since digital cameras can work under various lighting conditions. However, the existing photogrammetry approaches are not designed and are less suited for field conditions.

**OBJECTIVE AND SCOPE**

The objective of this paper is to develop a convenient and cost-effective procedure for the 3D reconstruction of individual aggregate particles from multi-view images. The proposed photogrammetry approach follows a marker-based design that enables background suppression, point cloud stitching, and scale referencing to obtain high-quality aggregate models. The approach should allow reconstruction across different aggregate sizes (especially for relatively large-sized aggregates) and is potentially extensible to work under field conditions. The equipment setup, reconstruction mechanism, and the key designs of the reconstruction approach are detailed in this paper. The accuracy of the reconstruction results is validated against ground-truth measurements for selected aggregate samples, and comparative analyses are conducted on 2D and 3D morphological properties to investigate the differences between the 2D and 3D statistics.

**MARKER-BASED 3D RECONSTRUCTION APPROACH**

**Equipment Setup**
The equipment of the reconstruction system includes a digital camera, a camera tripod, a 30.5-cm (12-in.) diameter turntable, and white cardboard background, as shown in **Figure 1**. The digital camera used in this study was a smartphone camera (Model: iPhone XR) with 4032-pixel by 3024-pixel resolution, but other types of digital cameras can also be used if the collected images are of sufficient quality and resolution. The camera was mounted on the tripod at a viewing angle of 30 degrees to 45 degrees with respect to the horizontal plane. A proper viewing angle ensures the top and side surfaces of the inspected aggregate particle are visible to the camera. During reconstruction, the camera was at a fixed position, and the multi-view images of the aggregate were obtained by manually rotating the turntable. The smartphone camera was programmed with an automatic shutter (with a beeping sound) every two seconds. In between two shutters, the operator rotates the turntable around 30 degrees and switches to the next view. Note that the use of a turntable and a white background with a fixed-position camera is one proposed setup to collect multiple views. The approach is flexible and designed to accommodate different configurations. For example, when applying this approach to larger aggregates that cannot easily fit onto a turntable, or a turntable is not available for field inspection, it is recommended to acquire multi-view images by moving the camera around the static object.

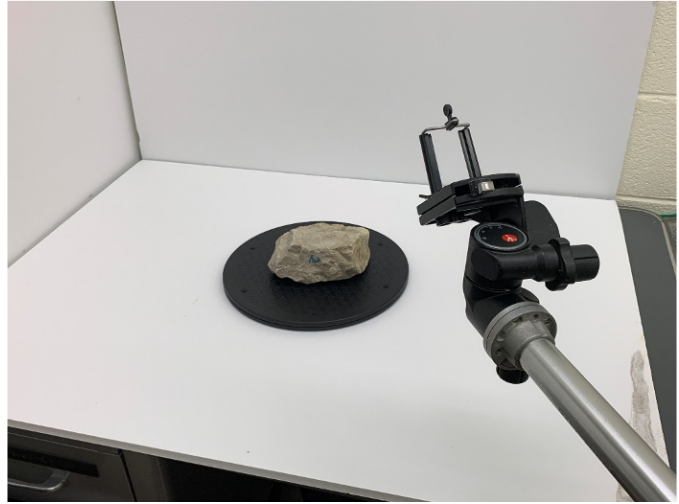

**Figure 1 Equipment setup for 3D reconstruction of aggregates.**

**Reconstruction by Structure-from-Motion**
In the computer vision domain, Structure-from-Motion (SfM) technique is a powerful photogrammetry method for 3D reconstruction of static scenes. The previous photogrammetry-based methods used by aggregate researchers (*22, 23*) also belong to the SfM category. SfM solves the problem of recovering 3D stationary structure from a collection of multi-view 2D images. A typical SfM pipeline involves three main stages: (a) extracting local features from 2D views and matching the common features across views, (b) estimating the motion of cameras and obtaining relative camera positions and orientations, and (c) recovering the 3D structure by jointly minimizing the total reprojection error (*24, 25*). The fundamentals and implementation of SfM are omitted from this discussion, but the key steps, (b) and (c), are discussed herein with necessary details. The process of simultaneously estimating the camera parameters and the 3D structure is also called bundle adjustment, which is essentially an optimization problem as shown in **Equation 1**:

$$\underset{\{P,X\}}{\text{Minimize}} \sum_{i=1}^{m} \sum_{j=1}^{n} \left\| x_{ij} - P_i X_j \right\|^2 \qquad (1)$$

where $P_i$ is the projection matrix of the $i^{th}$ camera, $X_j$ is the coordinates of the $j^{th}$ feature point in the 3D structure, and $x_{ij}$ is the projected pixel location of $X_j$ in the $i^{th}$ camera view. The total reprojection error, the objective function in **Equation 1**, is the squared pixel distance of all feature points across all camera views. Bundle adjustment process then iteratively finds the best estimates of the camera parameters and the point coordinates by minimizing the objective. After convergence, the reconstructed structure is available as a sparse 3D point cloud and can be further processed to generate a dense point cloud.

**Background Suppression by Masking for Noise Reduction**
The standard SfM procedure extracts features from the whole 2D images and attempts to reconstruct the entire scene, as shown in **Figure 2a**. This usually results in a 3D model that requires manual cleaning to remove unrelated background information (noise) and obtain a clean model of the aggregate sample only. Depending on how much of the background is reconstructed, the manual cleaning process could become considerably time-consuming, especially in regions where the aggregate is touching the background surface, as illustrated in **Figure 3a**. It is noteworthy that this manual cleaning requirement is not only limited to the SfM procedure. During the 3D reconstruction with costly devices (i.e., laser scanner, structured light scanner, etc.), manual cleaning is also a necessary step. This is because the scanning mechanism does not distinguish the foreground from background since their relative definition will vary from one application to another.

To reduce the various noise from unrelated background regions, the proposed approach improves the standard SfM approach by generating a foreground object mask $M$ for each image. During bundle adjustment, the object mask is applied as additional constraints in the original objective function, as shown in **Equation 2**:

$$\underset{\{P,X\}}{\text{Minimize}} \sum_{i=1}^{m} \sum_{j=1}^{n} M_{ij} \cdot \|x_{ij} - P_i X_j\|^2 \tag{2}$$

where $M_{ij}$ is the object mask indicating the inclusion or suppression of feature $X_j$ in the $i^{th}$ camera view.

The generation of this type of foreground object mask is an image segmentation problem. Although traditional segmentation methods can be applied using the color and edge information, the proposed approach adopts a deep learning-based segmentation method. The neural network architecture used is called U$^2$-Net, which is a successful design for the salient object detection task (26). Salient object detection is utilized to detect and extract the potential Region of Interest (RoI) of objects that may be salient in the image. The network uses deep nested U-shape convolutional-deconvolutional blocks to capture multi-scale contextual information without significantly increasing the computation cost. The training dataset was image-mask pairs prepared by both manual labeling and 3D to 2D projection of several manually cleaned 3D models. Based on experiments, around 100 image-mask pairs yield very robust and accurate foreground extraction for a given background environment. Note that for a given background environment, the network is trained only once, and no further training is involved in the reconstruction workflow. The raw images and the generated foreground masks of an example aggregate are illustrated in **Figure 2**.

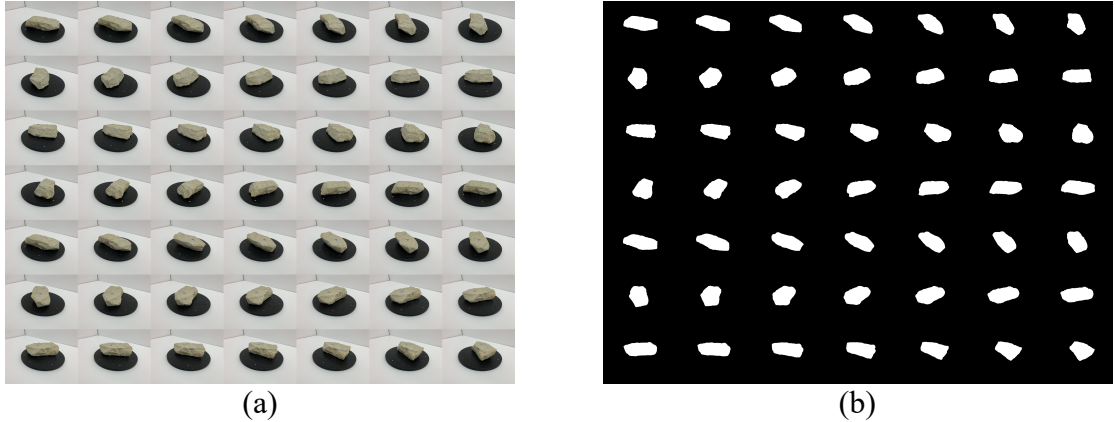

(a)                                           (b)

**Figure 2 (a) Multiple view images of an example aggregate particle, (b) salient object masks for each view.**

The reason of adopting a deep learning-based method is to improve the flexibility of the proposed approach. Although the experiments conducted in this paper were set up with a fixed background, the approach is designed to work in different environments, such as using different colors of the turntable and background, or under field conditions with natural lighting conditions. In such cases, a traditional segmentation method may not generate masks robustly; while the deep learning-based method only requires a few image-mask pairs to tune its behavior. The robustness of detection in natural background has been validated in the original U$^2$-Net development (*26*).

By applying the foreground masks, the unrelated background is suppressed, and the reconstructed model is noise-free and does not require any further manual cleaning. The resulting background suppression effect is illustrated in **Figure 3**.

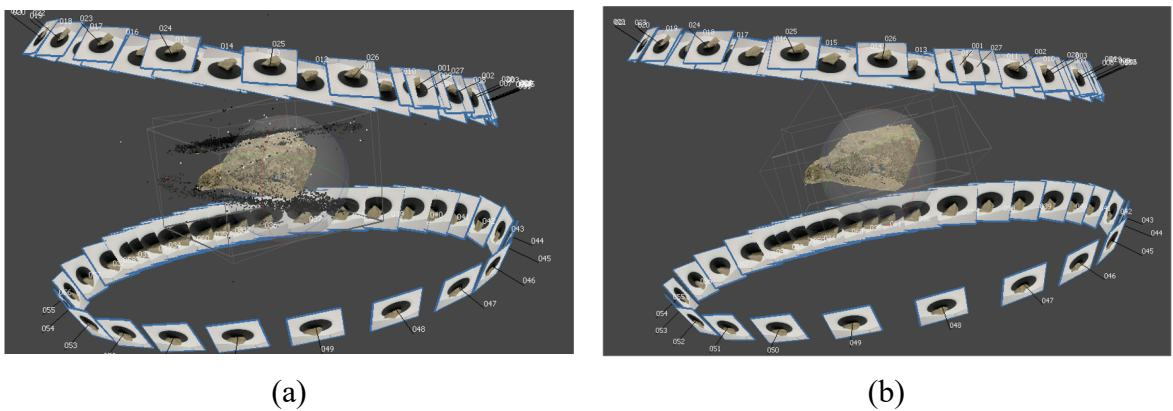

(a)                                           (b)

**Figure 3 Reconstructed sparse point cloud (a) without background suppression and (b) with background suppression.**

**Object Markers for Robust Point Cloud Stitching**
Unlike small-sized aggregates that can be easily elevated by a support pedestal, medium- and large-sized aggregates usually need to sit on a flat surface during reconstruction or scanning. This limits the possibility of obtaining all-around views of the aggregate and reconstructing with one run of SfM. Two or more rounds of reconstruction are required on different parts of the aggregate by adjusting its pose in between, and the partial point clouds must be stitched into a complete 3D model. The most common way to stitch multiple point clouds is to use point set registration algorithms (*27*). However, based on experiments, automatic registration algorithms are not always robust and may fail for certain aggregate samples with less distinct surface features.

In this regard, a set of object markers was designed to provide robust feature matching during point cloud stitching. Two markers were drawn with colored pencils on the side of each aggregate. The markers were designed to have a head-tail pattern with purple and red colors, as shown in **Figure 4a** and **Figure 4b**. Note that the selected colors are not fixed and can be adjusted based on the color of the aggregate for better contrast. The head and tail of each marker are the ends of a short and long line segments, respectively. Such pattern is invariant to different viewing angles and can thus be identified robustly. After the sparse reconstruction is completed, manual labeling of the markers is required on a few views (typically three views) to obtain a consistent localization of the markers in 3D coordinates. When the marker localization is completed for each partial point cloud, the stitching process can be conducted successfully, and a complete 3D model is obtained for the aggregate.

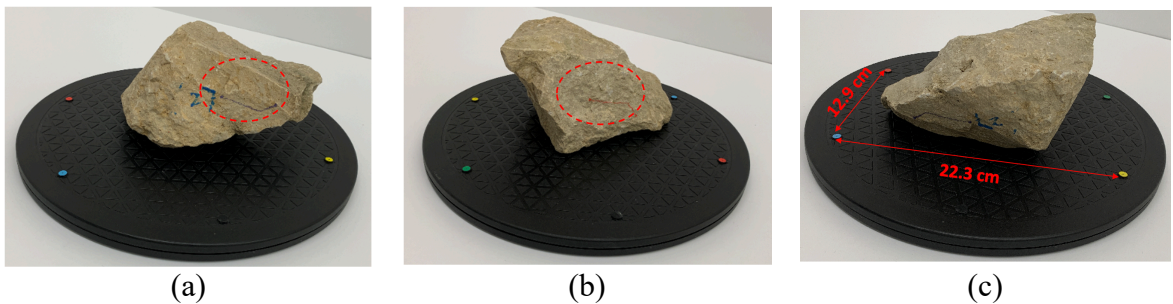

(a)          (b)          (c)

**Figure 4 (a) Purple-colored and (b) red-colored object markers for robust point cloud stitching, and (c) background markers for scale reference.**

**Background Markers for Scale Reference**
The reconstructed 3D model from previous steps is in a local coordinate system. To bring the model into a true physical scale and a global coordinate system, a set of background markers was designed to provide a scale reference. The design follows the same concept of Ground Control Points (GCP) in land surveying (*28*). Color-coding labels with red, green, blue, and yellow colors were placed at four corners of the turntable, as illustrated in **Figure 4c**. The distances between the markers were measured in advance and given as the scale factor. As discussed previously, when the proposed approach is applied without a turntable, the background markers could take other forms such as GCPs.

**Reconstruction Workflow and Results**
The reconstruction workflow can be summarized by the following steps:
- Step 1: *Preparation step* (executed only once for each environment). This involves setting up the equipment, tuning the foreground detection network, and placing the background markers.

- Step 2: *Placing the aggregate sample*. The sample is placed in the camera view, and object markers are labeled on the side surface.
- Step 3: *Capturing visible sides (two or more) of the sample*. By rotating the turntable (or moving the camera), multiple view images are taken. The same procedure is repeated for each side. In our experiments, 30 views were taken for each side with a two-second shutter interval, resulting in two minutes per sample for a two-side inspection.
- Step 4: *Reconstruction*. First, foreground masks are generated from the foreground detection network. Second, SfM is executed using the raw multi-view images and the associated foreground masks. Next, object markers and background markers are labeled on a subset of images (usually three images from each side). Finally, a complete 3D model is obtained by stitching the partial point clouds together.
- Steps 2 to 4 are repeated for each aggregate sample.

The reconstructed results presented in this paper were generated by extending the Agisoft Metashape (*29*) software program. Note that the implementation of the reconstruction step is not limited to certain software tools. Commercial software programs such as Agisoft Metashape (*29*), free software available such as VisualSFM (*30*), or open-sourced software available such as Meshroom (*31*), can all be extended to implement the proposed approach.

Example reconstruction results are visualized in **Figure 5**. The reconstructed model is available in different formats, such as the textured model that preserves the surface color information (**Figure 5a**), mesh model that shows the wireframe of vertex connectivity (**Figure 5b**), and point cloud model with discrete point coordinates (**Figure 5c**). In addition, an image collage of 40 aggregate samples reconstructed in this study is presented in **Figure 5d**. On average, each sample was exported at a resolution of around 50,000 vertices and 100,000 faces. It can be observed that the reconstructed aggregate models are of high quality and fidelity.

## MATERIAL INFORMATION AND GROUND-TRUTH VALIDATION

The outlined reconstruction procedure was used to inspect a set of 40 riprap aggregate particles collected from field visits to aggregate producers in Illinois. The samples conform to the 'RR3' category based on Illinois Department of Transportation (IDOT)'s specification (*32*), which typically refers to aggregates that have weights around 4.54 kg (10 lbs.). In the specification, RR1 and RR2 categories refer to small-sized riprap aggregates having the same size ranges as unbound base material in pavement engineering and ballast material in railway engineering, and RR3 to RR7 categories are medium to large-sized riprap aggregates that are more common in riprap applications. In terms of geological classification, these aggregate samples are dolomite rocks with a yellowish color, as shown in **Figure 5d**. Even though the results were demonstrated on relatively large-sized aggregates, the setup previously shown in **Figure 1** is expected to work for smaller sizes such as base course aggregates or ballast without further adjustments.

For each reconstructed aggregate particle, the basic 3D properties can be calculated from the 3D mesh model, including volume, surface area, and the shortest, intermediate, and longest dimensions in the three principal axes. The 3D properties of ten example aggregate particles are listed in **Table 1**. If the intermediate dimension is denoted as the nominal size of an aggregate, the size of these aggregate samples ranged from 7.6 cm (3 in.) to 15.2 cm (6 in.). For the ground-truth, the submerged volume of each aggregate sample was measured by a water displacement method following ASTM D6473 standard (*33*), listed as the measured volume in the second column in **Table 1**.

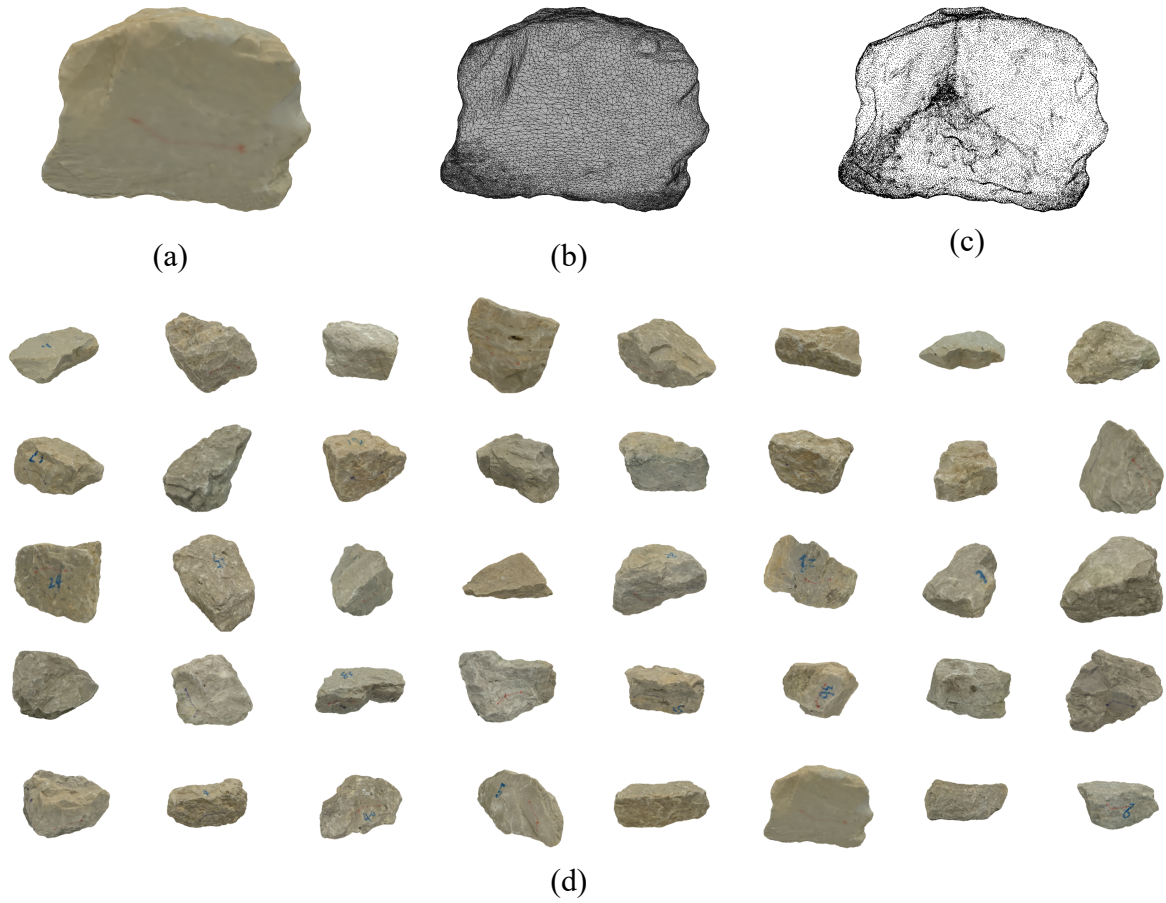

**Figure 5 (a) Textured model, (b) mesh model, (c) point cloud model of an example aggregate particle, and (d) collage of all reconstructed aggregate particles.**

**Table 1 Measured Volume, Reconstructed Volume, Area, and Principal Dimensions of Ten Example Particles**

| Rock ID | Measured Volume (cm³) | Reconstructed Volume (cm³) | Surface Area (cm²) | Shortest Dimension (cm) | Intermediate Dimension (cm) | Longest Dimension (cm) |
|---|---|---|---|---|---|---|
| 1 | 1014.9 | 1042.3 | 685.32 | 7.682 | 13.142 | 22.695 |
| 2 | 763.5 | 786.33 | 537.87 | 9.308 | 12.519 | 17.412 |
| 3 | 601.8 | 605.04 | 418.69 | 9.477 | 10.075 | 14.572 |
| 4 | 791.4 | 795.69 | 558.41 | 9.118 | 10.133 | 19.925 |
| 5 | 727.6 | 744.83 | 503.13 | 9.803 | 10.649 | 18.842 |
| 6 | 688.1 | 691.96 | 478.72 | 7.497 | 9.987 | 15.925 |
| 7 | 644 | 662.47 | 465.96 | 11.614 | 13.867 | 14.041 |
| 8 | 1140.5 | 1165.03 | 704.29 | 10.617 | 12.213 | 21.923 |
| 9 | 592.7 | 601.1 | 435.01 | 8.068 | 11.517 | 17.851 |
| 10 | 890.8 | 920.92 | 590.14 | 10.374 | 14.513 | 17.37 |

To validate the accuracy of the 3D reconstruction procedure, the reconstructed volume is compared against the measured ground-truth volume, as presented in **Figure 6**. A 45-degree line is plotted as a reference for the comparison. As the quantitative measure of accuracy, a statistical indicator, Mean-Percentage-Error (MPE), is calculated using **Equation 3**. Note that different from Mean-Absolute-Percentage-Error (MAPE), MPE can have a positive or negative sign, indicating a systematic overestimate or underestimate behavior, respectively.

$$MPE\ (\%) = \frac{\sum_{i=1}^{N} \frac{R_i - M_i}{M_i}}{N} \tag{3}$$

where
  $R_i$ is the reconstructed result of the $i^{th}$ sample,
  $M_i$ is the ground truth measurement of the $i^{th}$ sample, and
  $N$ is the total number of samples.

**Figure 6** shows a very good agreement between the reconstructed volume from the marker-based reconstruction approach and the ground-truth measured volume, with a MPE of +2.0%. The positive MPE also indicates a consistent, systematic overestimate of the reconstructed volumes. There are three potential reasons for this overestimation. First, the pixel locations of background markers are used to localize the marker in 3D coordinates. Therefore, pixel deviation when labeling the background markers may lead to a slight change of the scale reference. Second, a porous surface condition was observed on these dolomite aggregate particles, and the micro-texture areas that are filled with water during the measurement of the submerged volume may be reconstructed as flat faces. This could also lead to a systematic overestimate of the true submerged volume. Lastly, since SfM-based photogrammetry methods entail an optimization approach to jointly approximate the true object geometry, and cameras provide sparser representation (pixels) than laser scanning devices, it is reasonable to assume that certain systematic deviation may exist within acceptable accuracy.

When compared with the three-view reconstruction approach by Huang et al. (*14*), where a MAPE=5.1% (before averaging) and MAPE=3.6% (by averaging results from three repetitions) were obtained for the same aggregate particles, it is important to highlight the essential difference between the two approaches. First, the three-view reconstruction approach is a volumetric estimation approach rather than a true 3D reconstruction approach. The results from that approach are intersecting voxels (volume elements) that are simplified and approximated 3D representation of the sample, while the results from the approach developed in this study are true 3D mesh models of the sample. Second, the MAPE value with the three-view reconstruction approach is calculated after applying a complex volume correction step to the raw reconstructed volumes, while the volumes in this approach are raw volumes directly measured from the reconstructed models without any correction. Moreover, the three-view reconstruction approach targets quick size estimation in the field, yet the approach developed in this study focuses on the high-fidelity reconstruction of individual aggregates to obtain their true 3D models.

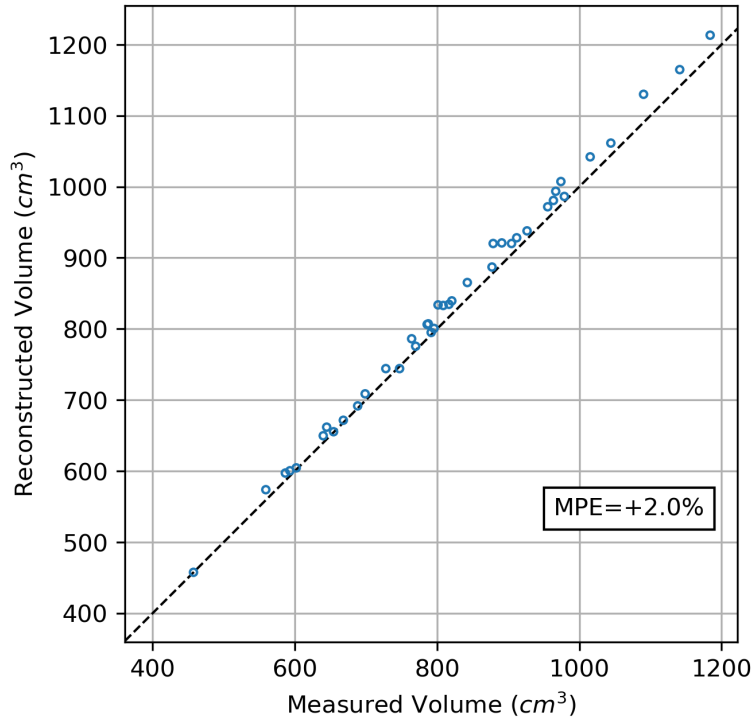

**Figure 6 Comparison of reconstructed volume and measured volume of aggregate samples.**

## COMPARATIVE ANALYSIS OF 2D AND 3D MORPHOLOGY OF AGGREGATES

Based on the multi-view images used during reconstruction and the resulting reconstructed 3D models, a comparative analysis is conducted to study the differences between 2D and 3D morphological properties of aggregates. The purpose of the comparative analysis is twofold: first, to check if major differences exist between 2D and 3D morphology indicators; second, to investigate the extent to which the morphological properties from 2D analysis can represent or indicate the true 3D morphological properties.

**Morphological Indicators for Comparative Analysis**
Since the comparative analysis is between 2D and 3D morphology, the morphological indicators should ideally have both the 3D version and its counterpart in 2D. Therefore, for the aspect ratio of particle shape, 2D and 3D Flat and Elongated Ratio (FER) are selected as the indicator pair, and for the roundness of particles, 2D circularity and 3D sphericity are selected as the indicator pair. The description of the morphological indicators is detailed herein.

**2D and 3D Flat and Elongated Ratios**
As the 2D indicator of particle aspect ratio, 2D FER is a widely used concept in both ASTM D4791 standard measurement and imaging-based approaches (*10, 11, 12, 34*). In image analysis, 2D FER is usually calculated from the particle silhouette after segmentation. Feret diameters (*35*) are measured along two perpendicular directions from different orientations. The maximum or longest Feret diameter, $L_{max}$, is obtained by searching for the longest edge-to-edge distance within the silhouette in all possible orientations, while the minimum or shortest Feret diameter, $L_{min}$, is obtained by searching for the shortest edge-to-edge distance within the silhouette that is perpendicular to the $L_{max}$ direction. The 2D FER is then defined as follows (**Equation 4**):

$$FER_{2D} = \frac{L_{max}}{L_{min}}, (L_{max} \geq L_{min}) \tag{4}$$

As the 3D counterpart of the aspect ratio indicator, 3D FER can be calculated after finding the minimum volume bounding box of the particle. O'Rourke (*36*) developed algorithms to find the minimal enclosing box of a point set. First, for each possible direction originated from the particle centroid, a 3D local coordinate frame is formed in the orthogonal searching directions. Next, for each orthogonal pair, the three edge-to-edge distances (3D Feret diameters) within the point set are calculated. The volume of the bounding box can then be computed, and the Feret diameters of the minimum volume bounding box are denoted as the shortest dimension $a$, intermediate dimension $b$, and longest dimension $c$. Accordingly, the orthogonal pair associated with the minimum volume bounding box represents the three principal axes of the particle. The 3D FER can then be defined based on the principal dimensions (**Equation 5**):

$$FER_{3D} = \frac{c}{a}, (c \geq b \geq a) \tag{5}$$

**2D Circularity and 3D Sphericity**
To compare the roundness of particles, a compactness measure of irregular shape is selected as the indicator, which takes the form of circularity in 2D and sphericity in 3D. Both indicators measure how closely a shape resembles a perfect circle in 2D or sphere in 3D, which serves as the unity with value 1.0.

Given the area $A$ and the perimeter $P$ of a 2D silhouette, 2D circularity can be calculated as shown in **Equation 6**. As a reference, an equilateral triangle has a circularity of 0.605 and a square has a circularity of 0.785, with higher values indicating the 2D shape is closer to a perfect circle.

$$Circularity_{2D} = \frac{4\pi A}{P^2} \tag{6}$$

For 3D sphericity, Wadell (*37*) defined the sphericity as the ratio between the surface area of an equivalent sphere having the same volume as the particle, $S_e$, and the measured surface area of the particle, $S$. This is often called the true sphericity. Given the surface area $A$ and the volume $V$ of a 3D model, the 3D sphericity can be computed using **Equation 7**. As a reference, a tetrahedron has a sphericity of 0.67 and a cube has a sphericity of 0.81, again with higher values indicating the 3D shape is closer to a perfect sphere.

$$Sphericity_{3D} = \frac{S_e}{S} = \frac{\sqrt[3]{36\pi V^2}}{A} \tag{7}$$

Note that although the circularity and sphericity are considered counterparts in 2D and 3D, the 2D and 3D versions of a shape may not necessarily have the same value. For example, if we consider cube is a 3D version of the 2D square, its 3D sphericity (0.81) differs slightly from the 2D circularity (0.785). When comparing 2D circularity values with 3D sphericity values, this intrinsic difference should be recognized. However, the overall trend from angular shape to round shape is consistent for sphericity and circularity. With the morphological indicators introduced in this section, 2D and 3D morphology statistics can be compared quantitatively.

**Results and Discussion**
For 2D morphology, 2D FER and 2D circularity are calculated from the multi-view images used in the reconstruction. The average value from multiple views is reported with range and standard deviation. Since for each aggregate particle, its 2D statistic is a distribution covering values from multiple views, directly comparing the standard deviation among different samples is not valid because each sample may have different averages. Therefore, the Coefficient of Variance (CoV), i.e., ratio between the standard deviation and the average, is used to characterize the variation of each sample. For 3D morphology, 3D FER and 3D sphericity are calculated from the reconstructed 3D model. **Figure 7** presents the comparison between 2D and 3D morphology. Note that the horizontal axis is sorted based on the 3D statistics to better illustrate the trend.

     **Figure 7a** shows that 3D FER of an aggregate is consistently higher than the average 2D FER from multi-view images. The 3D FERs of the samples range from 1.0 to 3.0 with around 75% of the samples having 3D FER less than 2.0. As for average 2D FERs, the values range from 1.0 to 2.0 with more than 75% of the samples having an average 2D FER of less than 1.5. In addition to the average 2D FERs, the range bars illustrate the minimum and maximum 2D FER values across all multi-view images. It can be observed that the minimum 2D FERs usually reach 1.0 and the maximum 2D FERs have values close to the 3D FERs. This indicates that during multi-view 2D analysis, there are certain views that can better represent the true 3D FER than others. However, it should be stressed that 2D aggregate analysis is often times limited to a single-view analysis, e.g., in practical scenarios such as analyzing the aggregate shape during conveying process, from top-views of aggregates spread on a table, or one angled face in aggregate stockpiles (*10, 11, 12, 13, 15, 16, 38*). Therefore, the average 2D FER could represent the value that is most likely to be obtained from a single-view analysis.

     **Figure 7b** shows that the 3D sphericity of an aggregate is also consistently higher than the average 2D circularity from multi-view images. The 3D sphericities of the samples range between 0.70 and 0.85, and the average 2D circularities mostly lie between 0.65 and 0.80. Again, the range bars illustrate the minimum and maximum circularities across all multi-view images. Like the FER comparison, the maximum circularities may approach the 3D sphericities for several samples, but the average 2D circularities can be considered as the most common value that can be obtained from single-view analysis.

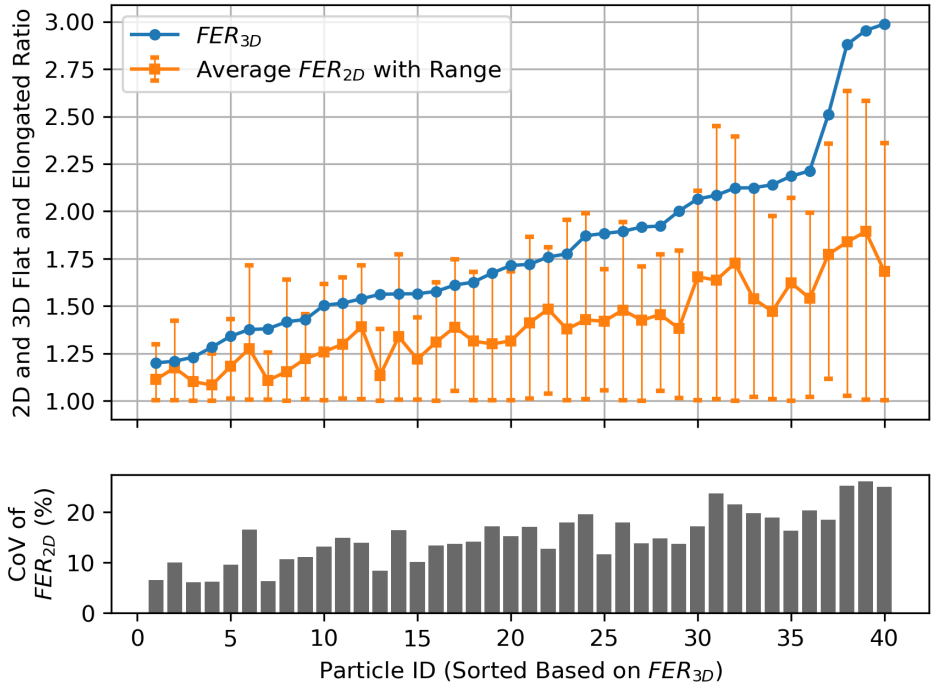

(a)

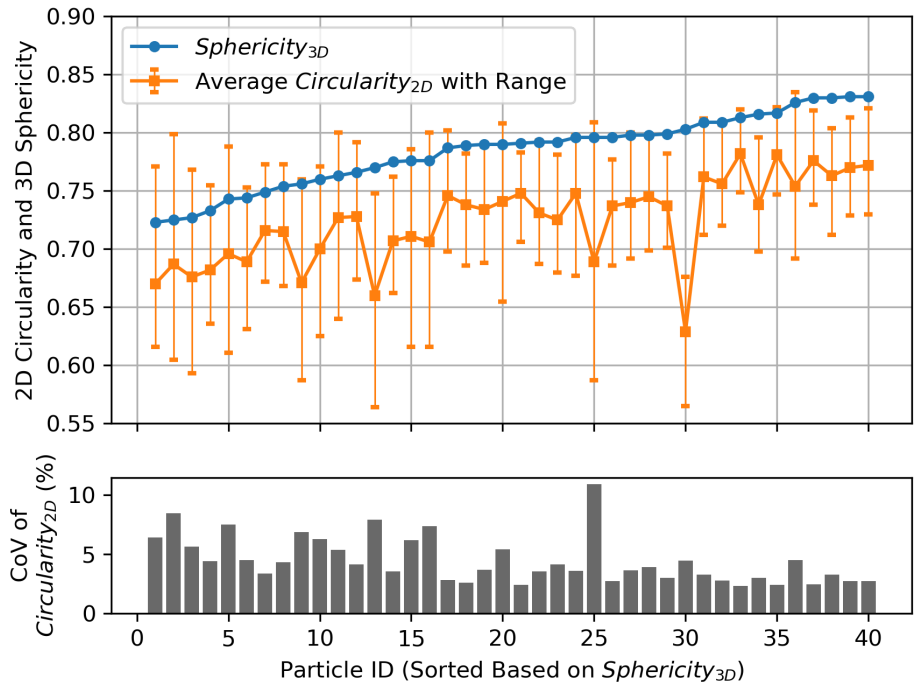

(b)

**Figure 7 (a) Comparison of 3D FER and 2D FER from multiple views, and (b) comparison of 3D sphericity and 2D circularity from multiple views.**

In **Figure 7a** and **Figure 7b**, the CoV chart is also presented below the main graph. The CoVs of 2D FER among all samples are mostly ranging between 10% and 20%, while the CoVs of 2D circularity show less variation with most values less than 10%. This may imply that the circularity indicator is usually less sensitive to varying views when compared to the FER indicator. Moreover, an opposite trend is observed in the FER CoV and circularity CoV: as the particle 3D FER increases, the CoV of 2D FERs tends to increase; as the particle 3D sphericities increases, the CoV of circularity tends to decrease. This is because higher 3D FER and lower 3D sphericity both indicate a 3D shape that is close to a rounded (less angular) sphere. Such uniform 3D shape results in less variance when projected into 2D silhouettes during multi-view analysis.

From **Figure 7** one can observe a consistent difference between the 2D and 3D morphologies, and the extent to which the morphological properties from 2D analysis can represent or indicate the 3D morphological properties needs to be further investigated. **Figure 8** illustrates the comparison between different ratios calculated from the 3D principal dimensions and the average 2D FER. In addition to the longest-to-shortest ratio (3D FER), the longest-to-intermediate (c/b) and intermediate-to-shortest (b/a) ratios are also plotted. Note that the c/b and b/a ratios must be both lower than the c/a ratio, but one ratio is not expected to be always lower or higher than the other, depending on the magnitude of the three principal dimensions. It is observed that in most cases, the average 2D FER falls within the envelope formed by the c/b and b/a ratios. This implies that the 2D FERs obtained from a single-view analysis are likely to capture the intermediate ratios among 3D principal dimensions rather than the true 3D FER. This also explains why the volume estimation step in pure 2D image analysis (such as I-RIPRAP, *16*) requires a 3D FER value to be given as an assumption. A correction factor could be applied to estimate the 3D FER from 2D FER upon further investigation using a comprehensive database of aggregate shapes.

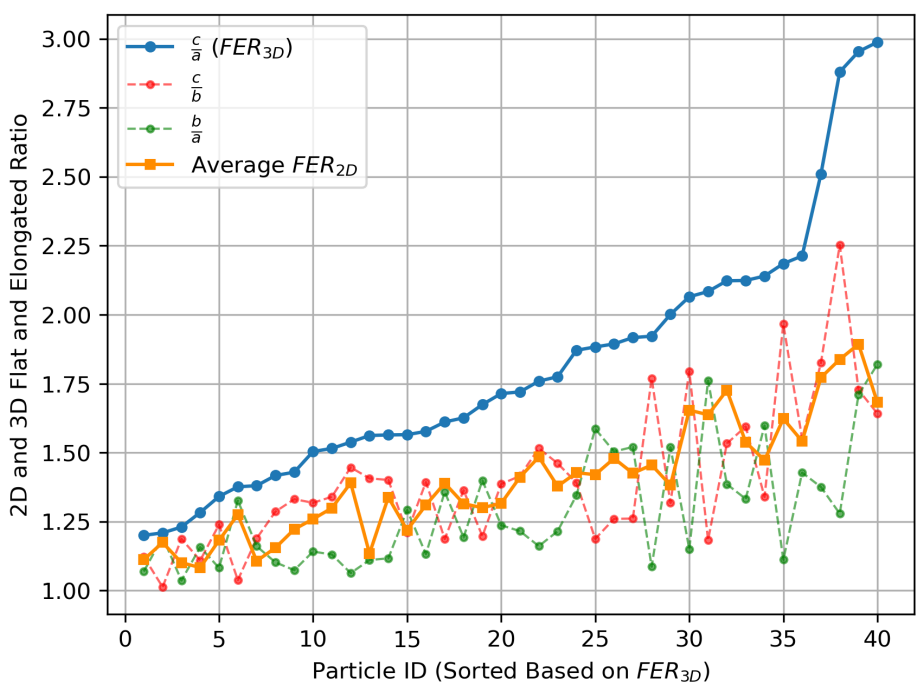

**Figure 8 Comparison of average 2D FER and the ratios of 3D principal dimensions.**

**CONCLUSIONS AND RECOMMENDATIONS**

The following conclusions can be drawn related to the proposed 3D reconstruction approach and the comparative analysis on 2D-3D morphologies:

- The literature review indicated that existing approaches for obtaining full 3D models of aggregates usually require using costly devices such as 3D structured light scanner, 3D laser scanner, and X-Ray Computed Tomography equipment. Several photogrammetry-based approaches also exist but are mostly limited to smaller-sized aggregates.
- A marker-based 3D reconstruction approach was developed as a cost-effective and flexible procedure to allow full reconstruction of aggregates. The proposed approach is a photogrammetry-based method with auxiliary designs to achieve background suppression, robust point cloud stitching, and scale reference. The approach was demonstrated on relatively large-sized aggregates, and the reconstructed models showed good agreements with ground-truth measurements.
- Comparative analysis was conducted between the 2D morphological properties from multi-view images and the 3D morphological properties from the reconstructed aggregate models. Significant differences were observed between the 2D and 3D statistics, which suggests that 2D morphological properties must be used carefully to infer the true 3D properties.

The proposed 3D reconstruction approach demonstrates great potential for serving as a flexible and cost-effective technique for aggregate inspection and data collection. By tuning the background suppression step to account for natural backgrounds, the approach can be easily extended for aggregate reconstruction in field scenarios. For quality assurance and quality control tasks, the approach provides useful 3D morphological information for aggregate quality assessment. For researchers, the approach can be used to collect data to establish an aggregate particle library, which is in great demand for simulation studies such as Finite Element Modeling (FEM) and Discrete Element Modeling (DEM). Moreover, from the comparison of 2D and 3D morphologies, further research and an in-depth investigation on different aggregate size groups is recommended to determine whether using a correction factor is applicable between the 2D and 3D morphology indicators.

**ACKNOWLEDGEMENTS**

This publication is based on the results of ICT-R27-214 project titled "3D Image Analysis Using Deep Learning for Size and Shape Characterization of Stockpile Riprap Aggregates." ICT-R27-214 was conducted in cooperation with the Illinois Center for Transportation; the Illinois Department of Transportation, Office of Program Development; and the U.S. Department of Transportation, Federal Highway Administration. The authors would like to acknowledge the members of IDOT Technical Review Panel (TRP) for their useful advice at different stages of this research. The help and support of Sheila Beshears of RiverStone Group, Andrew Buck and Daniel Barnstable of Vulcan Materials with the aggregate sample collection is greatly appreciated. The contents of this paper reflect the views of the authors who are solely responsible for the facts and the accuracy of the data presented. This paper does not constitute a standard, specification, or regulation.

AUTHOR CONTRIBUTION STATEMENT

The authors confirm contribution to the paper as follows: study conception and design: Erol Tutumluer, Haohang Huang, Jiayi Luo, John M. Hart, Issam Qamhia, and Andrew Stolba; data collection: Haohang Huang, Jiayi Luo, and John M. Hart; analysis and interpretation of results: Haohang Huang, Jiayi Luo, Issam Qamhia, and John M. Hart; draft manuscript preparation: Haohang Huang, Jiayi Luo, Issam Qamhia, John M. Hart, Erol Tutumluer, and Andrew Stolba. All authors reviewed the results and approved the final version of the manuscript.